\documentclass{article}

\usepackage{PRIMEarxiv}

\usepackage[utf8]{inputenc} 
\usepackage[T1]{fontenc}    
\usepackage{hyperref}       
\usepackage{url}            
\usepackage{booktabs}       
\usepackage{amsfonts}       
\usepackage{nicefrac}       
\usepackage{microtype}      
\usepackage{lipsum}
\usepackage{fancyhdr}       
\usepackage{graphicx}       
\usepackage{subfigure}
\usepackage{algorithm}
\usepackage{algorithm}
\usepackage{algorithmicx}
\usepackage{algpseudocode}
\usepackage{float}
\usepackage{multirow}
\graphicspath{{media/}}     

\title{Reinforced Pedestrian Attribute Recognition with Group Optimization Reward
}

\author{
  Zhong Ji, Zhenfei Hu, Yaodong Wang \\
  School of Electrical and Information Engineering \\
  Tianjin University \\
  Tianjin\\
  \texttt{\{jizhong, hzf0226, wangyaodong@tju.edu.cn\}tju.edu.cn} \\
   \And
  Shengjia Li \\
  R\&D Department \\
  China Academy of Launch Vehicle Technology \\
  Beijing\\
  \texttt{sjli@tju.edu.cn} \\
}

\begin{document}
\maketitle

\begin{abstract}
Pedestrian Attribute Recognition (PAR) is a challenging task in intelligent video surveillance. Two key challenges in PAR include complex alignment relations between images and attributes, and imbalanced data distribution. Existing approaches usually formulate PAR as a recognition task. Different from them, this paper addresses it as a decision-making task via a reinforcement learning framework. Specifically, PAR is formulated as a Markov decision process (MDP) by designing ingenious states, action space, reward function and state transition. To alleviate the inter-attribute imbalance problem, we apply an Attribute Grouping Strategy (AGS) by dividing all attributes into subgroups according to their region and category information. Then we employ an agent to recognize each group of attributes, which is trained with Deep Q-learning algorithm. We also propose a Group Optimization Reward (GOR) function to alleviate the intra-attribute imbalance problem. Experimental results on the three benchmark datasets of PETA, RAP and PA100K illustrate the effectiveness and competitiveness of the proposed approach and demonstrate that the application of reinforcement learning to PAR is a valuable research direction.
\end{abstract}


\section{Introduction}
\begin{figure}[t]
	\begin{center}
		\includegraphics[width=13cm]{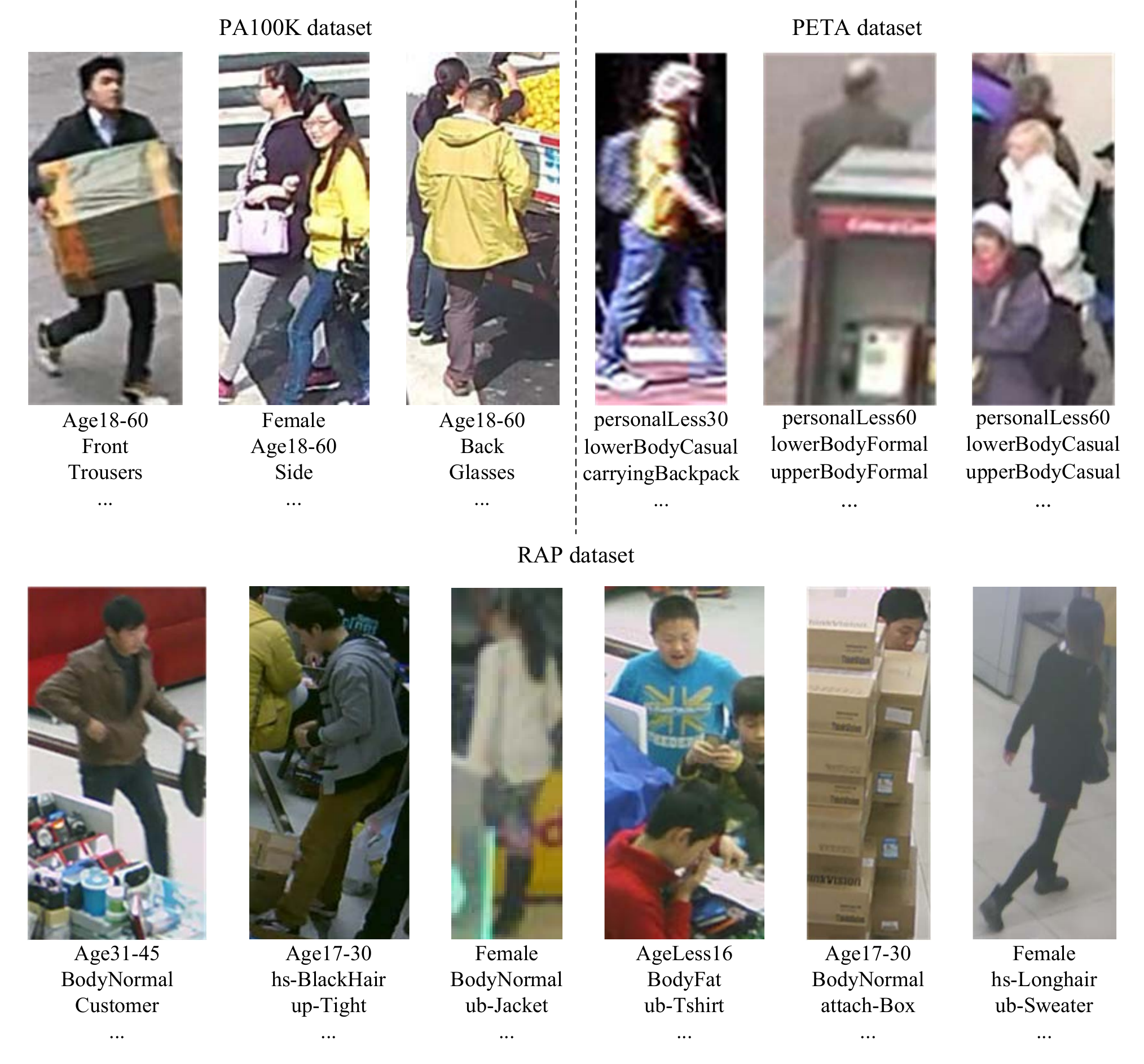}
	\end{center}
	
	\caption{Example images and the corresponding attributes.}
\label{fig1}
\end{figure}

Nowadays, intelligent video surveillance technology has been widely deployed \cite{yi2021robust,chen2022multiscale}. Pedestrian attributes, such as age, clothing style, gender, and accessary, are important soft-biometrics in video surveillance applications, such as person re-identification \cite{xu2021,liu2022feature,wang2022refining,pan2022aagcn}, person search \cite{ji2020multimodal,zhang2019efficient}, human parsing \cite{9535214}, and pedestrian detection \cite{ding2021robust,tang2022multi}. Thus, the recognition of them, called Pedestrian Attribute Recognition (PAR), has received great attention in recent years. Fig. \ref{fig1} shows some examples from the popular PETA \cite{deng2014pedestrian} , RAP \cite{li2016richly} and PA100K \cite{liu2017hydraplus} datasets.

Generally, PAR has two key challenges: complex alignment relations between images and attributes, and imbalanced data distribution \cite{ji2020pedestrian}. Since most pedestrian images are captured at far distance from real surveillance scenarios, they always show ambiguous appearance due to occlusion, low image resolution, and illumination. Meanwhile, the pedestrian attributes are quite diverse, for example, some of the “upperbody-Cotton” are in long style while some are in short; some of the “upperbody-Cotton” are zippered, some are not. Thus, the appearance ambiguity and attribute diversity make it quite hard to align the images and attributes. Some approaches employed attention-based sequence to sequence model to explore the complex alignment relations. For example, JRL \cite{wang2017attribute} applies recurrent attention mechanism on the output of encoder and reformulates the attribute decoding algorithm to focus on relevant parts when predicting the current attribute.  

\begin{figure*}[htb]
 
	\centering     
	\quad
	\subfigure[RAP dataset]{
	\includegraphics[width=16cm]{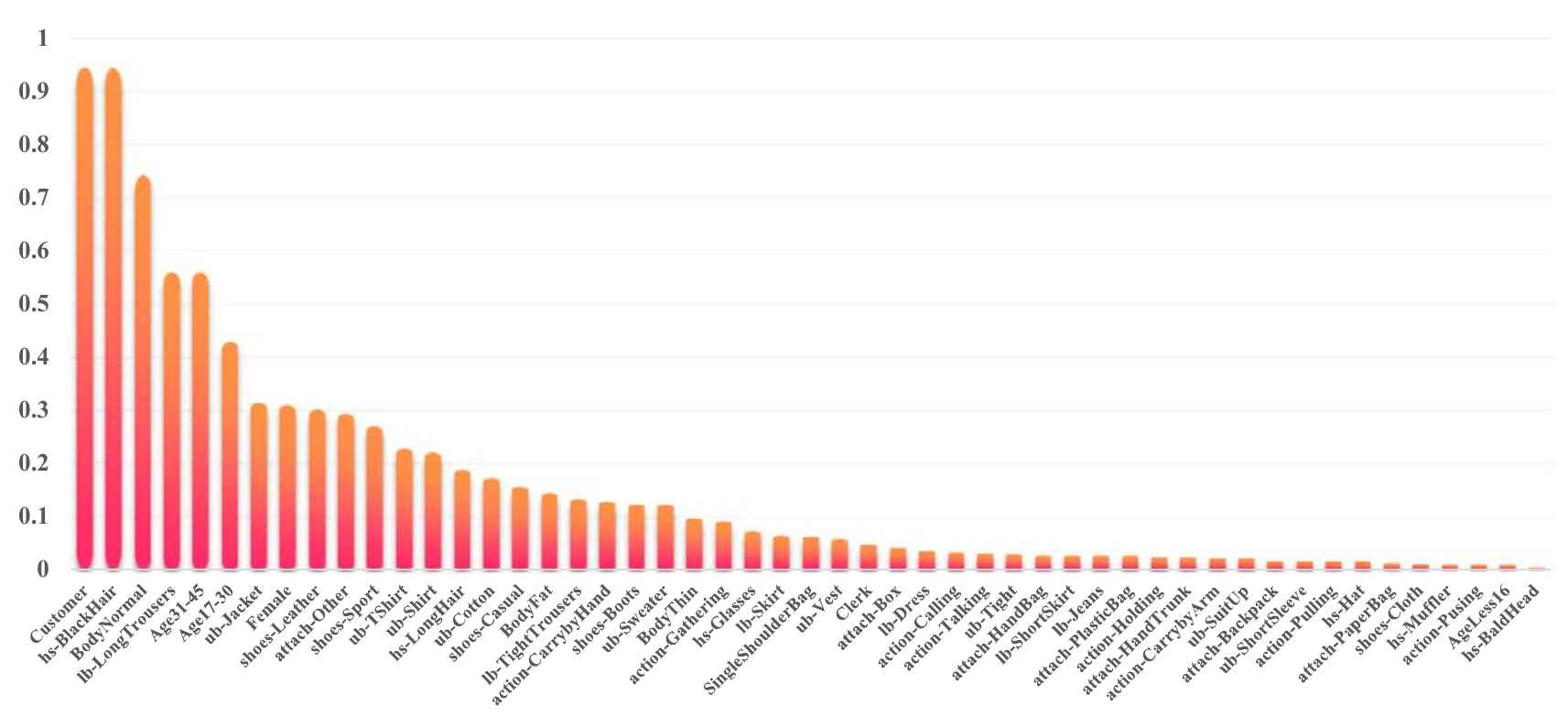}
	}
	\quad
	\subfigure[PA100K dataset]{
	\includegraphics[width=16cm]{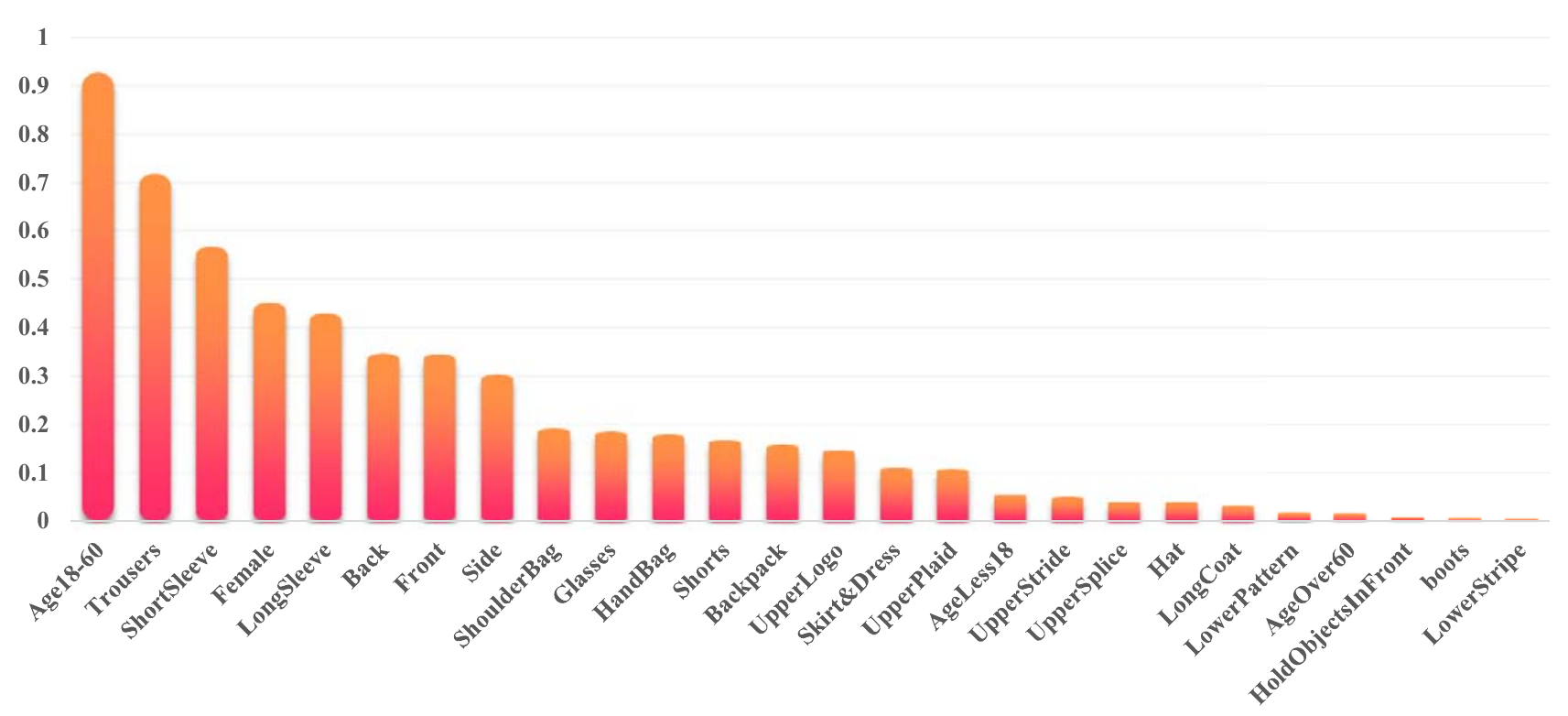}
	}
	\caption{The imbalanced attribute distributions of the RAP and PA100K datasets. }
\label{fig2} 
\end{figure*}

The second challenge is the imbalanced data distribution, which has always been an active research topic in computer vision filed. In PAR, this imbalance is manifested in two aspects, i.e., inter-attribute and intra-attribute imbalance. On the one hand, the inter-attribute distributions are imbalanced, that is to say, some attributes have a large proportion, while some are extremely few. Fig. \ref{fig2} illustrates the imbalanced attribute distributions of the RAP and PA100K datasets. We could observe, for example, that the ratio of “hs-BlackHair” is 94.48\%, while that of “hs-BaldHead” is only 0.38\% in RAP dataset. Some approaches alleviate this type imbalance problem by designing special loss functions. For instance, $\mathrm{IA}^{2}$-Net \cite{ji2019image} proposes a Focal Cross-Entropy (FCE) loss by combining the cross-entropy loss and the focal loss. Further, MTA-Net \cite{ji2020pedestrian} proposes a Focal Balanced Loss (FBL) on the basis of FCE loss by increasing the cost of difficult-to-identify attributes. On the other hand, the presence or absence for a single attribute can be regarded as an intra-attribute imbalanced distribution. For example, the presence probability of “hs-Hat” is only 1.67\%, while its absence probability is as high as 98.33\%. For a specific attribute, such imbalanced distribution enforces the network pay more attention to the majority samples, while ignoring the other valuable few samples. However, this type of imbalance problem is rarely tackled in existing PAR approaches.

Interestingly, almost all the existing approaches regard PAR as a recognition task. Inspired by a large number of effective applications of reinforcement learning (RL) in computer vision, we address PAR with RL by regarding PAR as a decision-making process, that is, a determination whether an attribute exists. Actually, RL has attracted increasing attentions as the success of AlphaGo  in 2015 \cite{silver2016mastering}. Afterwards, it has been successfully applied to many directions in computer vision filed, such as video summarization \cite{zhou2018deep}, pedestrian tracking \cite{zhang2021visual}, and image classification \cite{lin2020deep,he2018reinforced}. For example, in \cite{lin2020deep}, ICMDP formulates binary imbalance classification problem as a sequential decision-making process, and solve it by Deep Q-leanring. He $et\, al.$ \cite{he2018reinforced} introduced Deep Q-learning to curriculum learning for multi-label image classification, which explored the influence of label prediction order on the prediction results. 

However, directly employing existing RL-based approaches in PAR has the following two limitations. First, it is hard to be applied to a dataset with too many labels for a single image. This is because as the number of labels increases, the corresponding action space will increasing, which will aggravate the overestimation generated by Deep Q-learning \cite{van2016deep}. Secondly, the existing RL approaches cannot make use of the relationship among pedestrian attributes, nor can they effectively alleviate the imbalanced data distribution in PAR datasets. 

To address the above challenges and limitations, we propose a novel group reinforcement learning framework. Moreover, we apply an attribute grouping strategy and a group optimization reward function to alleviate the imbalance data distribution problem.

The contributions are highlighted as follows:

\begin{itemize}
\item We address the PAR task in a Deep Q-learning framework, as shown in Fig. 3. Although PAR is an attractive paradigm, it has rarely previously been exploited in a deep reinforcement learning framework. Particularly, we define PAR as a Markov decision process by designing image features and encoded attribute labels as states, utilizing binary codes 0 and 1 as action space, and giving corresponding positive or negative rewards according to whether the prediction result is accepted. In this way, the complex alignment between the images and attributes are modeled implicitly. To our best knowledge, it is the first work that defines PAR as a Markov decision process. 
\item We apply an Attribute Grouping Strategy (AGS) based on the correlation among the attributes in the same pedestrian image region to alleviate the inter-attribute imbalance problem. Further, we propose a Group Optimization Reward (GOR) function to alleviate the intra-attribute imbalance problem, in which the reward function of each group is optimized according to the overall situation of each group of attributes.
\item Extensive experiments on three benchmark datasets, i.e. PETA, RAP and PA100K, demonstrate the effectiveness and competitiveness of our proposed Rein-PAR approach.
\end{itemize}

The remaining sections of the paper are organized as follows. Section 2 reviews the related work. Section 3 introduces our proposed Rein-PAR in detail. Section 4 presents the experiments and ablation study, followed by the conclusion in Section 5.

\section{Related work}
Pedestrian attributes recognition has been broadly studied for many years. The applications of reinforcement learning in the field of computer vision are also received great attention. This section briefly introduces some works related to these two directions.

\subsection{Pedestrian Attribute Recognition}
Early PAR approaches applied hand-crafted features, such as Histogram of Oriented Gradients (HoG) \cite{sharma2011learning} and texture histogram \cite{layne2012person,layne2012towards}. With the renaissance of neural networks, deep learning based approaches have dominated the current PAR approaches. Generally, they can be divided into four groups, that are global-based, part-based, attention-based, and loss-based approaches.

\textbf{Global-based approaches:} As one of the pioneering deep PAR approaches, DeepMAR \cite{li2015multi} takes the whole image as input and presents a weighted cross-entropy loss to handle the attribute imbalance challenge. GSR-MAR \cite{siadari2019gsr} converts low-resolution images into high-resolution images through the Global Super Resolution Network for PAR. Li $et\, al.$ \cite{li2017sequential} proposed  a CNN-RNN  based sequential prediction model that takes global images as input, which can effectively encode scene context and  inter-person social relations. JLAC \cite{tan2020relation} employs graph convolution network to explore the relationship between attributes, which effectively improves the performance of attribute recognition.

\textbf{Part-based approaches:} As PAR is actually a fine-grained recognition task, PGDM \cite{li2018pose} recognizes pedestrian attributes by exploring human structure knowledge. It employs pre-trained pose estimation model to obtain keypoints of human image, and extracts part regions according to the keypoints for attribute recognition. Depend on the class activation maps, Liu $et\, al.$ \cite{liu2018localization} proposed a Localization Guide Network (LGNet), which captures the activation box for each attribute by cropping the high response area of the corresponding activation map. DTM+AWK \cite{zhang2020deep} leverages pose keypoints as auxiliary information to assist in positioning the attribute region. Tang $et\, al.$ \cite{tang2019improving} proposed  an Attribute localization module (ALM) which can discover the most discriminative regions adaptively. The application of auxiliary information has been proven to be effective, however, they \cite{li2018pose,liu2018localization,zhang2020deep} largely depend on the accuracy of positioning.

\textbf{Attention-based approaches:}  HP-Net \cite{liu2017hydraplus} applies the attention mechanism to PAR, which consists of two modules: Main Net (M-net) and Attentive Feature Net (AF-net). The AF-net contains multiple branches of multi-directional attention modules, which are applied to different semantic feature levels. In \cite{sarafianos2018deep}, DIAA introduces a multi-scale attention mechanism by directing the network to pay more attention to the spatial parts containing relevant information of the input image. Wu $et\, al.$ \cite{wu2020distraction} proposed a coarse-to-fine attention mechanism to reduce the irrelevant interference areas, which effectively improves the discriminant ability of attribute recognition.  JRL \cite{wang2017attribute} utilizes an encoder-decoder architecture to process image context and attribute correlation and applies attention mechanism to better focus on the local regions and obtain more accurate representation. CAS-SAL-FR \cite{yang2021cascaded} proposes a cascaded Split-and-Aggregate Learning (SAL) that captures the individuality and commonality of all attributes simultaneously at the feature map level and feature vector level with designed attribute-specific attention module (ASAM) and constrained losses. Although the visual attention mechanism has been successfully applied to PAR, the complexity of pedestrian images makes attention masks fail to obtain the position of a specific attribute. 

\textbf{Loss-based approaches:} The design of innovative loss function to alleviate data imbalance in PAR task is a hot research direction. DeepMAR \cite{li2015multi} proposes a weighted sigmoid cross entropy loss on the basis of the sigmoid cross entropy loss. Even though some attributes occupy a large proportion, they are still difficult to identify. Therefore, MTA-Net \cite{ji2020pedestrian} proposes Focal Balance Loss (FBL) based on FCE Loss \cite{ji2019image} by increasing the cost of those attributes difficult to recognize.

Different from the existing approaches that formulate PAR as a recognition process, our proposed Rein-PAR approach is a decision-making process, which formulates PAR in the reinforcement learning framework.

\subsection{Reinforcement Learning in Computer Vision}

The purpose of RL is to learn a policy for the agent from experimental trials by maximizing expected future rewards. It has been demonstrated to be effective in many computer vision tasks \cite{le2021deep}. These approaches can be divided into two categories, one is partial-RL approaches and the other is full-RL approaches.

\textbf{Partial-RL approaches:} This line of approach refers to that RL is applied to deal with part of the problem in a task. For example, RL-RBN \cite{wang2020mitigating} applies RL successfully to learn an adaptive margin policy to mitigate bias among different races and learn more balanced features, thus effectively improving the racial fairness in face recognition task. In action recognition task, Dong $et\, al.$ \cite{dong2019attention} found that irrelevant frames have an adverse impact on action prediction, and then applied RL and attention mechanisms to seek the most discriminate frames. In person Re-ID task, AAB \cite{zhang2020person} employs RL to design a novel Attribute Selection Module (ASM) to discard the noisy attributes and select the key ones. In person retrieval task, APN \cite{shi2020adaptive} utilizes policy gradient algorithm to optimize the agent to dynamically generate the optimal partitioning strategies for different images, which effectively reduces the human intervention problem. Wang $et\, al.$ \cite{Wang_2020_CVPR} modeled the online key decision process in dynamic video segmentation as a deep RL problem, and obtained an efficient and effective scheduling policy by leveraging expert information and appropriate training strategies. Zhang $et\, al.$ \cite{zhang2021visual} proposed a hierarchical RL framework for visual tracking called PACNet, which consists of the Policy and Actor-Critic Networks.  

\textbf{Full-RL approaches:} They formulate a certain task as a Markov decision process in an RL framework, which is more challenging than Partial-RL approaches. For instance, Zhou $et\, al.$ \cite{zhou2018deep} formulated video summarization as a sequential decision making process, and proposed a label-free reward function that jointly explain the diversity and representativeness of generated summaries. Guo $et\, al.$ \cite{guo2018dual} designed a deep RL based approach, which formulates the object tracking problem as a Markov decision process where state consisted of image regions extracted by the bounding box and eight actions. ICMDP \cite{lin2020deep} models the binary classification problem as a Markov decision process, and applies the Deep Q-learning algorithm to train the agent. He $et\, al.$ \cite{he2018reinforced} combined Deep Q-learning with curriculum learning to enable the agent to sequentially predict labels based on image feature and previously predicted labels. In \cite{bueno2017hierarchical}, the authors casted the objecte detection problem as a Markov decision 
process. The agent is utilized to find a region of interest in the image first, and then reduce the region of interest to find the smaller region based on the previously selected region, thus forming a hierarchy. Our Rein-PAR falls into this group, which formulates PAR as a Markov decision process in a Deep Q-leanring framework.

\begin{figure*}[t]
	\tiny
	\begin{center}
		\includegraphics[height=13.5cm]{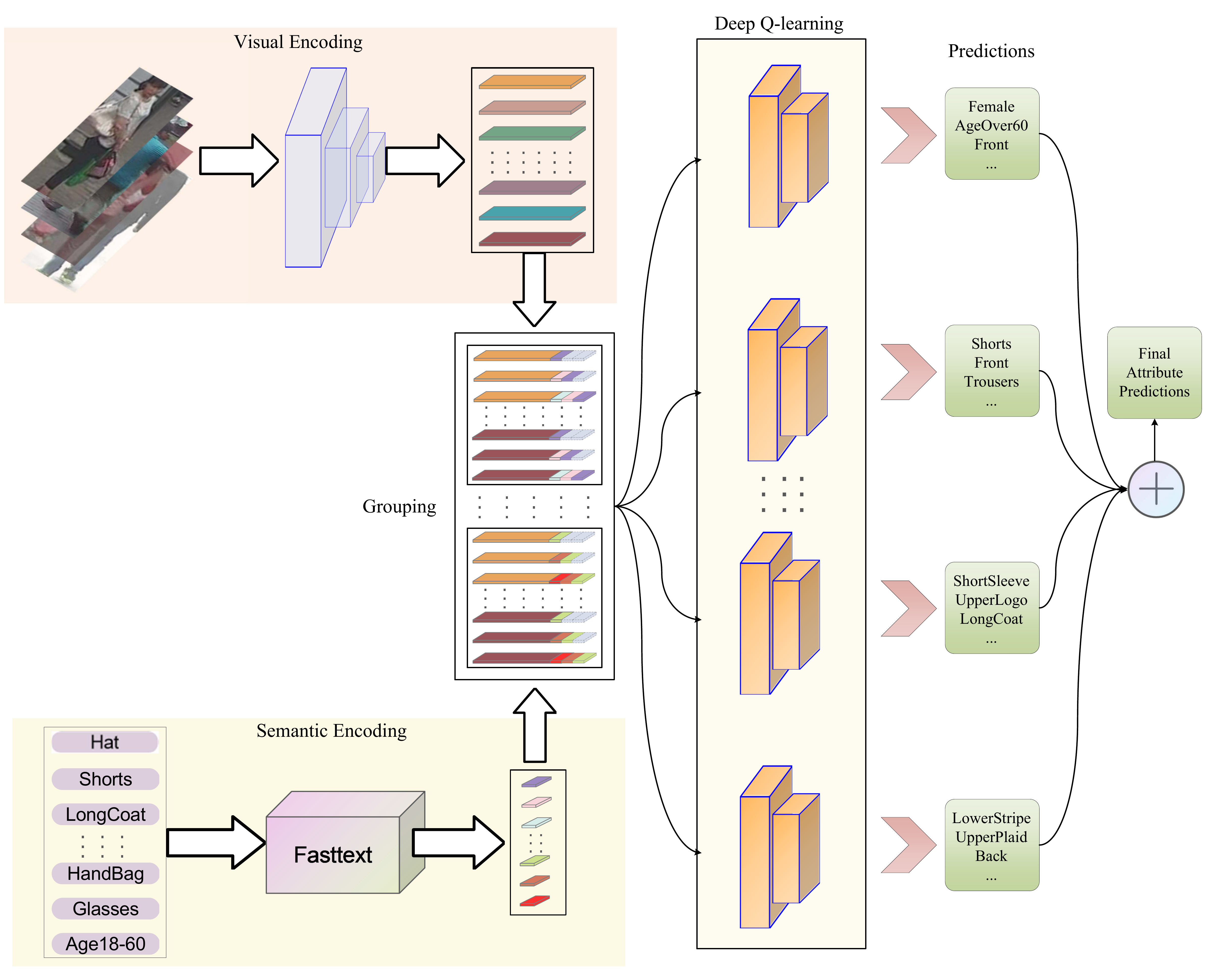}
	\end{center}
	
	\caption{Architecture of our proposed Rein-PAR model.}
\label{fig3}
\end{figure*}

\section{Method}
Fig. \ref{fig3} is the overall framework of our Rein-PAR. In this section, we first introduce in detail how we formulate PAR as a Markov decision process. Then, we present how to train the agent using Deep Q-learning \cite{mnih2015human}. After that, the grouping of attributes is detailedly described. Finally, we introduce our proposed group optimization reward function.
\subsection{PAR as Markov Decision Process}

We formulate PAR as a Markov decision process, which can be addressed with reinforcement learning algorithms. The agent is used to determine whether attributes exist, whose goal is to learn a
policy to maximize the total discount reward, a larger
discount reward means a higher the accuracy of recognition. We define the feedbacks from the environment as reward to the agent, and regard the recognition process of an image as an episode. 

Markov decision process can be described by a five-tuple $(S, A, R, \mathcal{T}, \gamma)$, where $S$ is the state space, $A$ is a limited set of actions, $R$ represents the reward brought by the transition from state $s$ to state $s^{\prime}$ via action $a$, $\mathcal{T}$ represents the transition from the previous state to the next state, and $\gamma$ is a discount factor. Their details are as follows.

\textbf{State:}  State refers to the information of the observed environment. In practical situations, the information that can be observed  is very limited due to cognitive limitations. In our approach, we represent it as a tuple, including image features and attribute information, expressed as :
\begin{figure*}[pt]
	\begin{center}
		\includegraphics[height=6.9cm]{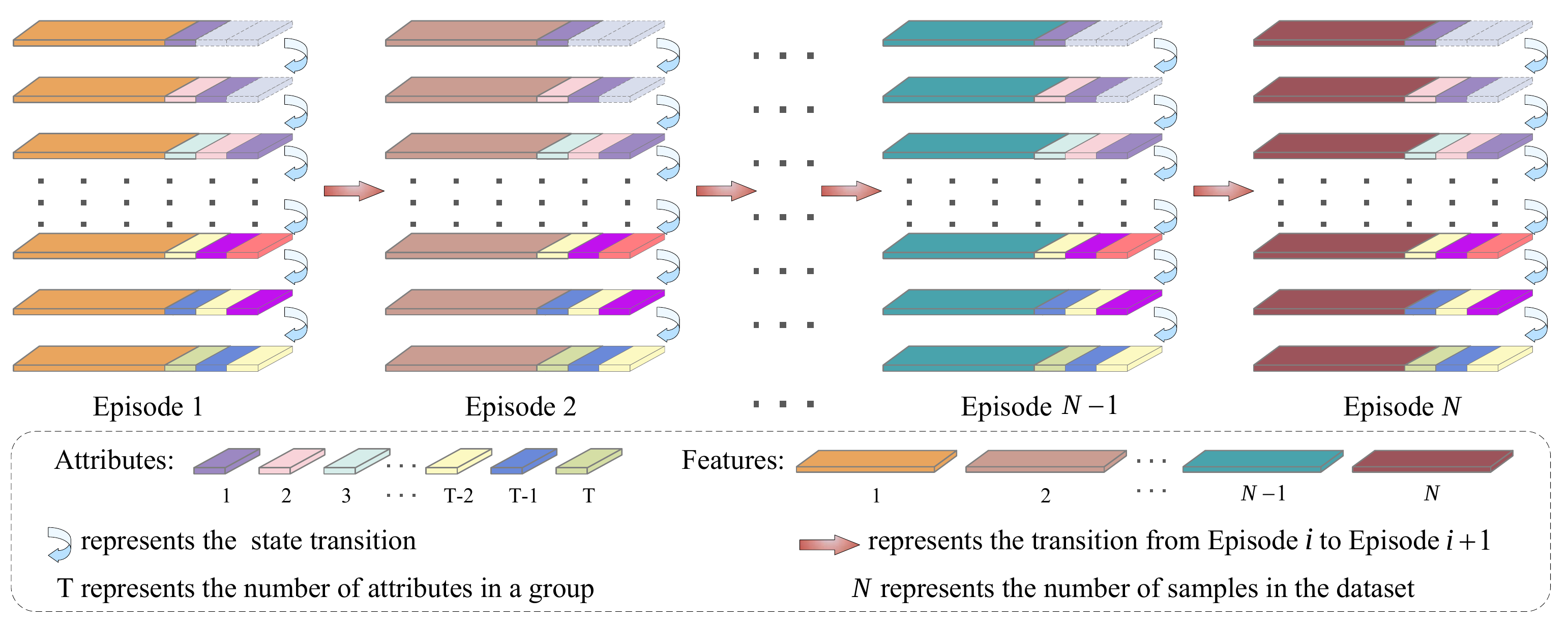}
	\end{center}
	
	\caption{The state transition process of Rein-PAR. The concept of "group" will be presented in the attribute grouping strategy.}
\label{fig4}
\end{figure*}
\begin{equation}
	s=(f, v),
\end{equation}
where $s \in S$ summarizes the information of the observed image and the attribute,  $f$ is the feature of the current image, and $v$ is the attribute vector. The 2048-dimensional image features are extracted by the poplar Resnet-50 model  \cite{he2016deep}, which is pre-trained model on ImageNet \cite{russakovsky2015imagenet}. Since the object categories in ImageNet are different from those in PAR task, we fine-tuned it on each PAR training dataset.

The attribute information refers to the attributes to be predicted at the current time $t$ and those have been predicted at the time $t-1$ and $t-2$. It is represented by a fasttext-encoded vector \cite{joulin2016bag}. For each attribute, we encode it into an $L$-dimensional vector, where $L$ is the number of attributes in the dataset.

\textbf{Action:} Action refers to the behavior issued by the agent and the interaction between the agent and the environment. RMIC \cite{he2018reinforced} made an attempt by directly utilizing label set as the action space, which is likely to cause overestimation problem. Usually, a smaller action space avoids the overestimation problem. By doing so, we define the agent's action space as a set $A=\{0,1\}$, which determines whether the corresponding attribute exists. Particularly, the element 0 represents that there is no corresponding attribute, and that of 1 represents the existence of the corresponding attribute. In the training stage, the agent interacts with the environment after taking action, and obtains the corresponding positive or negative rewards feedback from the environment. For each image, the number of actions depends on the number of labels, and all actions corresponding to each image are the final predicted labels of the image.

\textbf{Reward:} Reward is a significant research direction in RL, which refers to the value that an agent receives after conducting an action to describe whether its behavior is good or bad. In our approach, a basic reward could be set as $+1$ and $-1$, that is, when the value of the action is the same as the label, a positive reward is given, otherwise, a negative reward is given. It is expressed as:

\begin{equation}
	r= \left\{\begin{array}{cl} +1 & \mbox { if } \quad a_{t}=l_{t} \\ 
	      -1 & \mbox { if } \quad a_{t} \neq l_{t}\end{array}\right.,
\end{equation}
where $a_{t}$ is the action of the agent in state $s_{t}$, $l_{t}$ is the label of the attribute on the sample in state $s_{t}$.

However, this design ignores the intra-attribute imbalance problem, an elaborate reward will be introduced in the fourth part of this section.

\textbf{Transitions:} After an action is performed, the current state is transformed to the next state. In our approach, the MDP transitions are deterministic, that is, for each state there is a specified next state,  the action has no affect on the next state. The specific state transfer formula is as follows:

\begin{equation}
	\mathcal{T}(s, a)=\mathcal{T}((f, v), a)=\left(f, v^{\prime}\right)=s^{\prime},
\end{equation}

The state transition process is shown in Fig. \ref{fig4}.

\textbf{Discount factor:} Discount factor $\gamma \in[0,1]$  is to balance the relationship between immediate reward and future reward. When its value approaches 0, the agent focuses more on short-term return, while more long-term return is considered when it is closed to 1.

\begin{table*}[!t]
\centering
\begin{minipage}[!t]{\textwidth}
  \renewcommand{\arraystretch}{1}
  \caption{The groups of the PETA dataset, which are obtained according to the position and characteristic of the attributes.}
  \label{table1}
  \centering
  \setlength{\tabcolsep}{5mm}{
  	\begin{tabular}{c|c}
	\hline
	Group       & Attribute                                             \\ \hline
	Age, Gender & personalLess30, personalLess45, personalMale, etc.       \\ \hline
	Head        & Hat, Muffler, Sunglasses, hairLong                    \\ \hline
	Upper Body  & Casual, Formal, Jacket, Logo, Plaid, ShortSleeve, etc.  \\\hline
	Lower Body  & Formal, Jeans, Shorts, ShortSkirt, Trousers, etc.       \\ \hline
	Footwear    & LeatherShoes, Sandals, Shoes, Sneaker                 \\ \hline
	Carrying    & Backpack, Other, MessengerBag, Nothing, PlasticBags   \\ \hline
	\end{tabular}}
  \end{minipage}
\\[12pt]
\begin{minipage}[!t]{\textwidth}
  \renewcommand{\arraystretch}{1}
  \caption{\upshape  The groups of the RAP dataset, which are obtained according to the position and characteristic of the attributes.}
  \label{table2}
  \centering
  \setlength{\tabcolsep}{5mm}{
  \begin{tabular}{c|c}
	\hline
	Group                                                                 & Attribute                                                                                                                        \\ \hline
	\begin{tabular}[c]{@{}c@{}}Age, Gender\\ Bodyshape, Role\end{tabular} & \begin{tabular}[c]{@{}c@{}}AgeLess16, Age17-30, Age31-45, Female, BodyFat,\\  BodyNormal, BodyThin, Customer, Clerk\end{tabular} \\ \hline
	Head                                                                  & BaldHead, LongHair, BlackHair, Hat, Glasses, Muffler                                                                             \\ \hline
	Upper Body                                                            & \begin{tabular}[c]{@{}c@{}}Shirt, Sweater, Vest, TShirt, Cotton, Jacket, \\ SuitUp, etc.\end{tabular}                            \\ \hline
	Lower Body                                                            & LongTrousers, Skirt, ShortSkirt, Dress, Jeans, etc.                                                                              \\ \hline
	Footwear                                                              & Leather, Sport, Boots, Cloth, Casual                                                                                             \\ \hline
	Accessory                                                             & Backpack, SingleShoulderBag, HandBag, Box, etc.                                                                                  \\ \hline
	Action                                                                & Calling, Talking, Gathering, Holding, Pushing, etc.                                                                              \\ \hline
	\end{tabular}}
  \end{minipage}
\\[12pt]
\begin{minipage}[!t]{\textwidth}
  \renewcommand{\arraystretch}{1}
  \caption{\upshape  The groups of the PA100K dataset, which are obtained according to the position and characteristic of the attributes.}
  \label{table3}
  \centering
  \setlength{\tabcolsep}{5mm}{
  \begin{tabular}{c|c}
	\hline
	Group            & Attribute                                                                  \\ \hline
	Age, Gender      & AgeOver60, Age18-60, AgeLess18, Female                                     \\ \hline
	Location, Head   & Front, Side, Back, Hat, Glasses                                            \\ \hline
	Upper Body       & ShortSleeve, LongSleeve, UpperStride, UpperLogo, etc.                        \\ \hline
	\begin{tabular}[c]{@{}c@{}}Lower Body,\\  Foot\end{tabular} & \begin{tabular}[c]{@{}c@{}}LowerStripe, LowerPattern, LongCoat, Trousers, \\ Shorts, Skirt\&Dress, boots\end{tabular} \\ \hline
	Accessory        & HandBag, ShoulderBag, Backpack, HoldObjectsInFront                                                                                           \\ \hline
	\end{tabular}}
  \end{minipage}
  \vspace{-0.1cm}
\end{table*}
\subsection{Attributes Grouping Strategy}
There are strong correlations among the pedestrian attributes, especially those in the same region are often mutually exclusive. For example, ``BoldHair" and ``LongHair" cannot appear at the same time. We apply the grouping 
idea to group attributes according to their position and characteristic. Then, we make a separate prediction for each group. Through grouping, the correlation among attributes in the same group is more fully utilized. Moreover, the grouping strategy enables the distribution of attributes in the same group relatively balanced, which alleviates the inter-imbalance attributes distribution problem to some extent. The attributes groups are shown in Tables \ref{table1}, \ref{table2} and \ref{table3}:

\subsection{Group Optimization Reward Function}

Reinforcement learning algorithms are usually sensitive to the reward function, which is one of the key parts in RL. The typical one shown in Eq. (2) treats all attributes equally, which does not meet the practical PAR requirement. We could observe from Fig. \ref{fig2} that there are only five attributes in RAP dataset and three ones in PA100K dataset are common (the presence ration larger than 0.5), while most attributes in the PAR are less common. Thus, the appearance of an attribute is actually an imbalanced data distribution problem. This intra-attribute imbalanced distribution enforces the training of the network biased towards the common attributes. To alleviate this problem, we propose a Group Optimization Reward (GOR) function to pay more attention to those less common attributes, which optimizes the reward function on the basis of grouping. It is designed as follows:

\begin{equation}
	r^{\prime}=\left\{\begin{array}{cc}
	1 & a_{t}=1 \& l_{t}=1 \\
	-1 & a_{t}=0 \& l_{t}=1 \\
	-\rho & a_{t}=1 \& l_{t}=0 \\
	\rho & a_{t}=0 \& l_{t}=0
	\end{array}\right.,
	\end{equation}

\begin{equation}
	\rho=\left\{\begin{array}{lc}
	0.15 & c \in[0,0.05) \\
	0.25 & c \in[0.05,0.25) \\
	0.35 & c \in[0.25,0.35) \\
	0.45 & c \in[0.35,0.45) \\
	0.55 & c \in[0.45,1)
	\end{array}\right.,
	\end{equation}

\begin{equation}
	c=\frac{\sum_{T} n_{T}}{T^{*} N-\sum_{T} n_{T}},
\end{equation}
where $c$ refers to the imbalanced coefficient, $T$ is the number of attributes in the group, $N$ is the size of the dataset, $n_{T}$ is the number of images containing a certain attribute,  $a_{t}$ is the action of the agent in state $s_{t}$, and $l_{t}$ is the label of the attribute on the sample in state $s_{t}$. It should be noticed that $c$ is the ratio of the sum of the attributes presence to the sum of the attributes absence in a group, which reflects the overall situation of each group of attributes.

It could be observed that the reward function is assigned a ``+1" or ``-1" feedback in case the attribute is present, and a $\rho$  or $-\rho$ feedback for the opposite case. In this way, the less common attributes generally receive a smaller feedback in the case of its absence in a sample. Since the absence of the less common attributes predominate in the samples, it alleviates the intra-attribute imbalance problem existing in most attributes. However, it should be pointed out that the reward function will have a negative influence on the recognition of the common attributes, but for the overall recognition result, its influence is positive.

\subsection{Deep Q-learning for Rein-PAR}
\begin{figure}[t]
	\begin{center}
		\includegraphics[width=7cm]{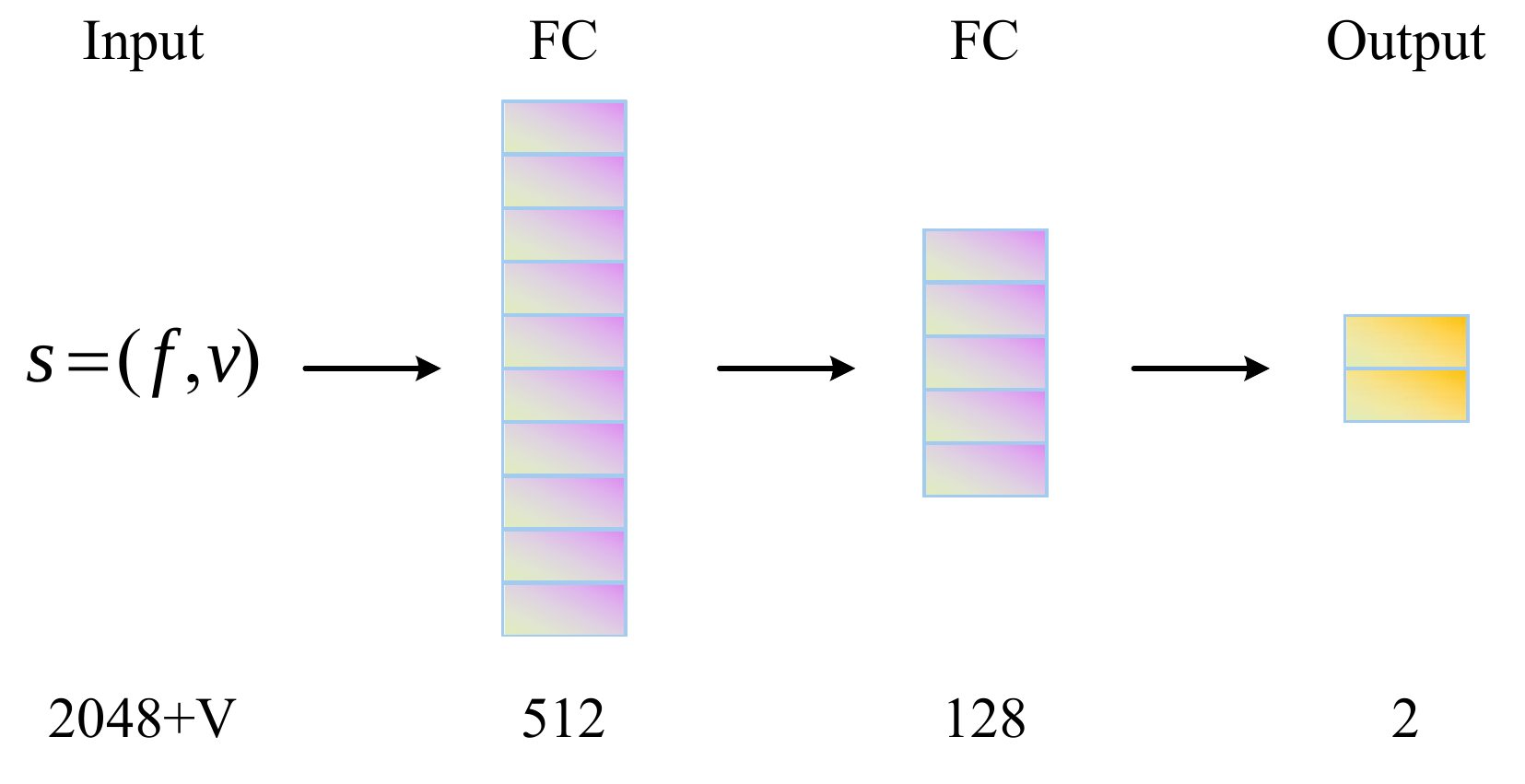}
	\end{center}
	
	\caption{The architecture of our Deep Q-learning network.}
\label{fig5}
\end{figure}

The optimal policy to maximize discount rewards can be obtained via RL. In Rein-PAR, we resort to the Deep Q-learning \cite{mnih2015human} algorithm to train our agent, as shown in Fig. \ref{fig5}. It consists of three fully connected layers, 512, 128 and 2, respectively. The input is the state composed of image features and attribute-encoded vectors. Our training goal is to enable the agent to recognize attributes as
accurately as possible.

During the training process, we store the transition information in the replay memory in the form of $\left(s, a, r, s^{\prime}\right)$ . It should be noted that the use of replay memory allows the agent to train without considering the order of images and attributes. When the policy network is updated, a mini batch randomly sampled from the replay memory is applied for training. The loss function is the (Mean Square Error) MSE loss as follow:

\begin{equation}
	L(\theta)=\mathbb{E}_{\left(s, a, r, s^{\prime}\right) \sim D}\left[\left(r+\hat{Q}\left(s^{\prime}, a^{\prime} ; \theta^{-}\right)-Q(s, a ; \theta)\right)^{2}\right],
\end{equation}
where $r+\hat{Q}\left(s^{\prime}, a^{\prime} ; \theta^{-}\right)$ is from the target network, $\theta^{-}$ is the parameter in the target network, $Q(s, a ; \theta)$ is the output value in the policy network, and  $\theta$ is the weights in the policy network. We use Adam optimization algorithm to update network parameters.

The Deep Q-learning \cite{mnih2015human} for Rein-PAR algorithm is shown as Algorithm 1. This process can be understood as the process of judging all attributes of each image as a game (episode), and each judging whether an attribute exists in an image is regarded as a step in the game, and the epoch refers to how many times a dataset is trained. The purpose of training is to enable the agent to judge all attributes of an image as correctly as possible.
\renewcommand{\algorithmicrequire}{\textbf{Initialization:}}  
\renewcommand{\algorithmicensure}{\textbf{Train:}} 

\begin{algorithm}[!htbp]
  \caption{Deep Q-learning for Rein-PAR} 
  \label{alg:Framwork}   
  \begin{algorithmic}[0]
    \Require
    \State Initialize Replay Memory $D$ to capacity $B$;
    \State Initialize the whole action space $A$;
    \State Initialize action-value function $Q$ with random weights $\theta$;
    \State Initialize  target action-value function $\hat Q$ with random weights ${\theta ^ - } = \theta $;
    \Ensure
       the training process for Rein-PAR;

       \For{epoch = 1,$M$}
         \For{each image }
      \State Initialize the current state $s_0$ with the current image 
      \State feature $f$ and semantic vector $v_i$;
      \label{ code:fram:extract }
            \For{t = 1,T}
            \State Select action $a_t$ from $A$ with $ \varepsilon-greedy$ policy
            \State Execute $a_t$, observe reward $r_t$, next state ${s_{t + 1}}$
            \State Store transition ($s_t$, $a_t$, $r_t$, ${s_{t + 1}}$) in D 
            \State Sample random minibatch of transitions
            \State ($s_j$, $a_j$, $r_j$, ${s_{j + 1}}$) from $D$
            \State
            \State Set ${y_j} \!= \!\left\{ \!{\begin{array}{*{20}{c}}
                \!{{r_j}\qquad\qquad\qquad{\rm{                            for\; terminal\; }}{s_{j + 1}}}\\
                \!{{r_j} \!+\! \gamma \mathop {\max }\limits_{a'} \hat Q\left( {{s_{j + 1}},{a^\prime };\!{\theta ^ - }} \!\right){\rm{     otherwise}}}
                \end{array}} \right.
                $
            \State Perform a gradient descent step on ${\left( {{y_j} - Q({s_{j + 1}},{a_j};\theta )} \right)^2}$ with respect to the network parameters $\theta$;
            
            \State  Update ${s_{t + 1}} \leftarrow {s_t}$ 
            \State Update $\hat Q = Q$ every $C$ step
     
     \EndFor
    \EndFor
    \EndFor
  \end{algorithmic}
\end{algorithm}

\section{Experiment}
We first briefly depict the experiment setup, including the datasets, implementation details and evaluation metrics. Then we show and analyze the experimental results. Finally, we conduct ablation experiments to analyze the impacts of attributes grouping strategy and group optimization reward function.
\subsection{Experiment Setup}

\textbf{Datasets:} We verify the effectiveness of the proposed approach on three benchmark datasets, namely PETA \cite{deng2014pedestrian}, PA100K \cite{liu2017hydraplus} and RAP \cite{li2016richly}. Concretely, PETA consists of 8,705 pedestrians and 19,000 images (resolution ranges from 17 $\times$ 39 to 169 $\times$ 365). The training, validation and test set of this dataset include 9,500, 1,900 and 760 images respectively. PA100K is the largest dataset used for PAR so far, which contains a total of 100,000 pedestrian images collected from outdoor surveillance cameras, each image has 26 popular attributes. The entire dataset is randomly divided into 80,000 training images, 10,000 validation images and 10,000 test images. RAP contains 41,585 pedestrian images from indoor scenes, of which 33,268 images are employed for training and 8,317 for testing.

\textbf{Implementation Details:} We train the network for 15 epochs, and set the capacity of replay memory $D$ to 2000, update frequency $C$ of the target network to 100 and the mini batch size to 64. The discount factor $\gamma$ is set to 0.9. The $\varepsilon \mbox {-greedy }$ strategy is utilized on action selection. During the training stage, the probabilities of exploration are gradually decreased from 0.9 to 0.05. During the test stage, we set the probability of exploration is to 0. The network is optimized by Adam optimizer and the default parameter Settings are adopted. Our approach is implemented on the publicly available Pytorch platform on a single NVIDIA GeForce GTX 1080Ti GPU with 12GB of memory.

\textbf{Evaluation Metrics:}  We evaluate the performance of our approach on five metrics, as shown in Eqs.(8)-(12). Among them, mean accuracy (mA) is a label-based metric, which treats each attribute independently. It first calculates the classification accuracy of positive and negative samples for each attribute, and then averages their values as the recognition result for the attributes. Accuracy (Acc), precision (Prec), recall rate (Rec) and F1 score are instance-based metrics. 

\begin{equation}
	m A=\frac{1}{2 L} \sum_{i=1}^{L}\left(\frac{T P_{i}}{P_{i}}+\frac{T N_{i}}{N_{i}}\right),
\end{equation}

\begin{equation}
	A C C=\frac{1}{N} \sum_{i=1}^{N} \frac{\left|Y_{i} \cap f\left(x_{i}\right)\right|}{\left|Y_{i} \cup f\left(x_{i}\right)\right|},
\end{equation}

\begin{equation}
	Prec=\frac{1}{2 N} \sum_{i=1}^{N} \frac{\left|Y_{i} \cap f\left(x_{i}\right)\right|}{\left|f\left(x_{i}\right)\right|},
\end{equation}

\begin{equation}
	Rec=\frac{1}{2 N} \sum_{i=1}^{N} \frac{\left|Y_{i} \cap f\left(x_{i}\right)\right|}{\left|Y_{i}\right|},
\end{equation}

\begin{equation}
	F1 = \frac{{2*\;Prec*Rec}}{{Prec + Rec}},
\end{equation}
where $L$ is the number of attributes, $N$ is the number of samples, $P_{i}$ and $N_{i}$ are the number of positive and negative samples, $TP_{i}$ and $TN_{i}$ are the number of correctly predicted positive and negative samples, $Y_{i}$ is the ground truth positive labels of the $i$-$th$ sample, and $f\left(x_{i}\right)$ is the predicted positive labels of the $i$-$th$ sample.

\subsection{Comparison with State-of-the-Art Methods}
\begin{table}[h]
	
	\caption{\upshape  Comparison results on PETA dataset. The best results are marked in boldface.}
	\centering
	\tabcolsep 15pt
	\renewcommand\arraystretch{1.2}
	\begin{tabular}{c|ccccc}
	\hline
	Method         & mA             & Acc  & Prec  & Rec            & F1    \\ \hline
	ACN \cite{sudowe2015person}           & 81.15          & 73.66 & 84.06 & 81.26          & 82.64 \\
	DeepMAR \cite{li2015multi}       & 82.89          & 75.07 & 83.68 & 83.14          & 83.41 \\
	SR \cite{liu2017semantic}            & 82.83          & -     & 82.54 & 82.76          & 82.65 \\
	CTX \cite{li2017sequential}           & 80.13          & -     & 79.68 & 80.24          & 79.68 \\
	WPAL \cite{zhou2017weakly}          & 85.50          & 76.98 & 84.07 & 85.78          & 84.90 \\
	HP-Net \cite{liu2017hydraplus}        & 81.77          & 76.13 & 84.92 & 83.24          & 84.07 \\
	VeSPA \cite{sarfraz2017deep}         & 83.45          & 77.73 & 86.18 & 84.81          & 85.49 \\ 
	PGDM \cite{li2018pose}          & 82.97          & 78.08 & 86.86 & 84.68          & 85.76 \\
	$\mathrm{IA}^{2}$-Net \cite{ji2019image}       & 84.13          & 78.62 & 85.73 & 86.07          & 85.88 \\
		JLPLS-PAA \cite{tan2019attention}          & 84.88          & 79.46 & 87.42 & 86.33          & 86.87 \\
	MT-CAS \cite{zeng2020multi}          & 83.17          & 78.78 & 87.49 & 85.35        & 86.41 \\
	MTA-Net \cite{ji2020pedestrian}       & 84.62          & 78.80 & 85.67 & 86.42          & 86.04 \\\hline
	Rein-PAR (Ours) & \textbf{85.51} & 78.45 & 84.08 & \textbf{88.77} & 85.91\\ \hline
	\end{tabular}
	\label{table4}
	\end{table}

\begin{table}[h]
	
	\caption{\upshape  Comparison results on RAP dataset. The best results are marked in boldface.}
	\centering
	\tabcolsep 15pt
	\renewcommand\arraystretch{1.2}
	\begin{tabular}{c|ccccc}
	\hline
	Method         & mA             & Acc  & Prec  & Rec            & F1    \\ \hline
	ACN \cite{sudowe2015person}           & 69.66          & 62.61 & 80.12 & 72.26          & 75.98 \\
	DeepMAR \cite{li2015multi}       & 73.79          & 62.02 & 74.92 & 76.21          & 75.56 \\
	CTX \cite{li2017sequential}           & 70.13          & -     & 71.03 & 71.20          & 70.23 \\
	JRL \cite{wang2017attribute}           & 77.81          & -     & 78.11 & 78.98          & 78.58 \\
	SR \cite{li2017sequential}            & 74.10          & -     & 75.11 & 76.52          & 75.83 \\
	WPAL \cite{zhou2017weakly}          & 81.25          & 50.30 & 57.17 & 78.39          & 66.12 \\
	HP-Net \cite{liu2017hydraplus}        & 76.12          & 65.39 & 77.33 & 78.79          & 78.05 \\
	VeSPA \cite{sarfraz2017deep}         & 77.70          & 67.35 & 79.51 & 79.67          & 79.59 \\
	PGDM \cite{li2018pose}          & 74.31          & 64.57 & 78.86 & 75.90          & 77.35 \\
	GSR-MAR \cite{siadari2019gsr}       & 67.76          & 63.44 & 82.27 & 71.82          & 76.69 \\
	$\mathrm{IA}^{2}$-Net \cite{ji2019image}       & 77.44          & 65.75 & 79.01 & 77.45          & 78.03 \\
	JLPLS-PAA \cite{tan2019attention}          & 81.25          & 67.91 & 78.56 & 81.45          & 79.98 \\
	MSE-Net \cite{lou2019mse}          & 71.12          & 62.43 & 78.82 & 72.94          & 75.77 \\
	
	MTA-Net \cite{ji2020pedestrian}       & 77.62          & 67.17 & 79.72 & 78.44          & 79.07 \\
	HR-Net \cite{an2020hierarchical}       & 81.10          & 45.70 & 51.48 & 78.56          & 62.20 \\
	VALA \cite{chen2022pedestrian}       & 78.33          & 67.48 & 79.81 & 80.84          & 80.32 \\
	\hline
	Rein-PAR (Ours) & \textbf{81.67} & 66.24 & 73.24 & \textbf{85.80} & 78.68 \\ \hline
	\end{tabular}
	\label{table5}
	\end{table}

\begin{table}[]
	
	\caption{\upshape  Comparison results on PA100K dataset. The best results are marked in boldface.}
	\centering
	\tabcolsep 15pt
	\renewcommand\arraystretch{1.2}
	\begin{tabular}{c|ccccc}
	\hline
	Method         & mA             & Acc   & Prec  & Rec            & F1    \\ \hline
	DeepMAR \cite{li2015multi}       & 72.70          & 70.39 & 82.24 & 80.42          & 81.32 \\
	HP-Net \cite{liu2017hydraplus}        & 74.21          & 72.19 & 82.97 & 82.09          & 82.53 \\
	VeSPA \cite{sarfraz2017deep}         & 76.32          & 73.00 & 84.33 & 81.49          & 83.20 \\
	PGDM \cite{li2018pose}         & 74.95          & 73.08 & 84.36 & 82.24          & 83.29 \\
	LGNet \cite{liu2018localization}         & 76.96          & 75.55 & 86.99 & 83.17          & 85.04 \\
	GSR-MAR \cite{siadari2019gsr}        & 72.43          & 73.46 & 87.68 & 79.94          & 83.63 \\ 
	MT-CAS \cite{zeng2020multi}          & 77.20          & 78.09 & 88.46 & 84.86        & 86.62 \\
	VALA \cite{chen2022pedestrian}       & 80.08          & 78.14 & 87.60 & 86.73          & 87.16 \\\hline
	Rein-PAR (Ours) & \textbf{80.55} & 77.20 & 84.76 & \textbf{87.67} & 85.70 \\ \hline
	\end{tabular}
	\label{table6}
	\end{table}

We choose eighteen state-of-the-art approaches for comparison, which include global-based models such as ACN \cite{sudowe2015person}, DeepMAR \cite{li2015multi}, GSR-MAR \cite{siadari2019gsr}, SR \cite{liu2017semantic}, CTX \cite{li2017sequential}, MSE-Net \cite{lou2019mse}, and HR-Net \cite{an2020hierarchical}; part-based models such as PGDM \cite{li2018pose} and LGNet \cite{liu2018localization}; attention-based models such as JRL \cite{wang2017attribute}, HP-Net \cite{liu2017hydraplus}, VeSPA \cite{sarfraz2017deep}, VALA \cite{chen2022pedestrian}, JLPLS-PAA \cite{tan2019attention}, and MT-CAS \cite{zeng2020multi} ; and loss function based models such as WPAL \cite{zhou2017weakly}, MTA-Net \cite{ji2020pedestrian}, and $\mathrm{IA}^{2}$-Net \cite{ji2019image}. The results on the three datasets are shown in Table \ref{table4}, Table \ref{table5} and Table \ref{table6}, respectively. We have the following observations:
\begin{itemize}
	\item[1)]
	It could be observed that our Rein-PAR achieves a competitive performance on the three datasets. It should be noted that most approaches reported their results on only two datasets due to the diverse data challenge, and it is really hard to achieve the best performance on all the five metrics.  
	\item[2)]
	Compared with the competitors, Rein-PAR is the best on the metrics of mA and Rec on the three datasets. For example, on the RAP dataset, the mA reaches 81.67\%, 0.42\% higher than the second-ranked approach WPAL, and 4.35\% higher than the second-ranked VeSPA approach on the Rec metric. The highest mA indicates that our approach is better than the competitors in the recognition accuracy of a single attribute. The highest Rec indicates that our approach has a higher accuracy in discriminating samples that are indeed positive. The main reason is that we apply GOR to pay more attention to those less common attributes.
	\item[3)]  
	Rein-PAR is also comparable with the competitors on the other metrics. Specifically, our approach outperforms most approaches on the metrics Acc and F1. Our approach is inferior on the Prec metric, whose reason lies in that our approach pays less attention to negative samples. This is the direction that we need to improve in our future work. 
\end{itemize}

	\begin{figure*}[pt]
		\begin{center}
			\includegraphics[height=9.5cm]{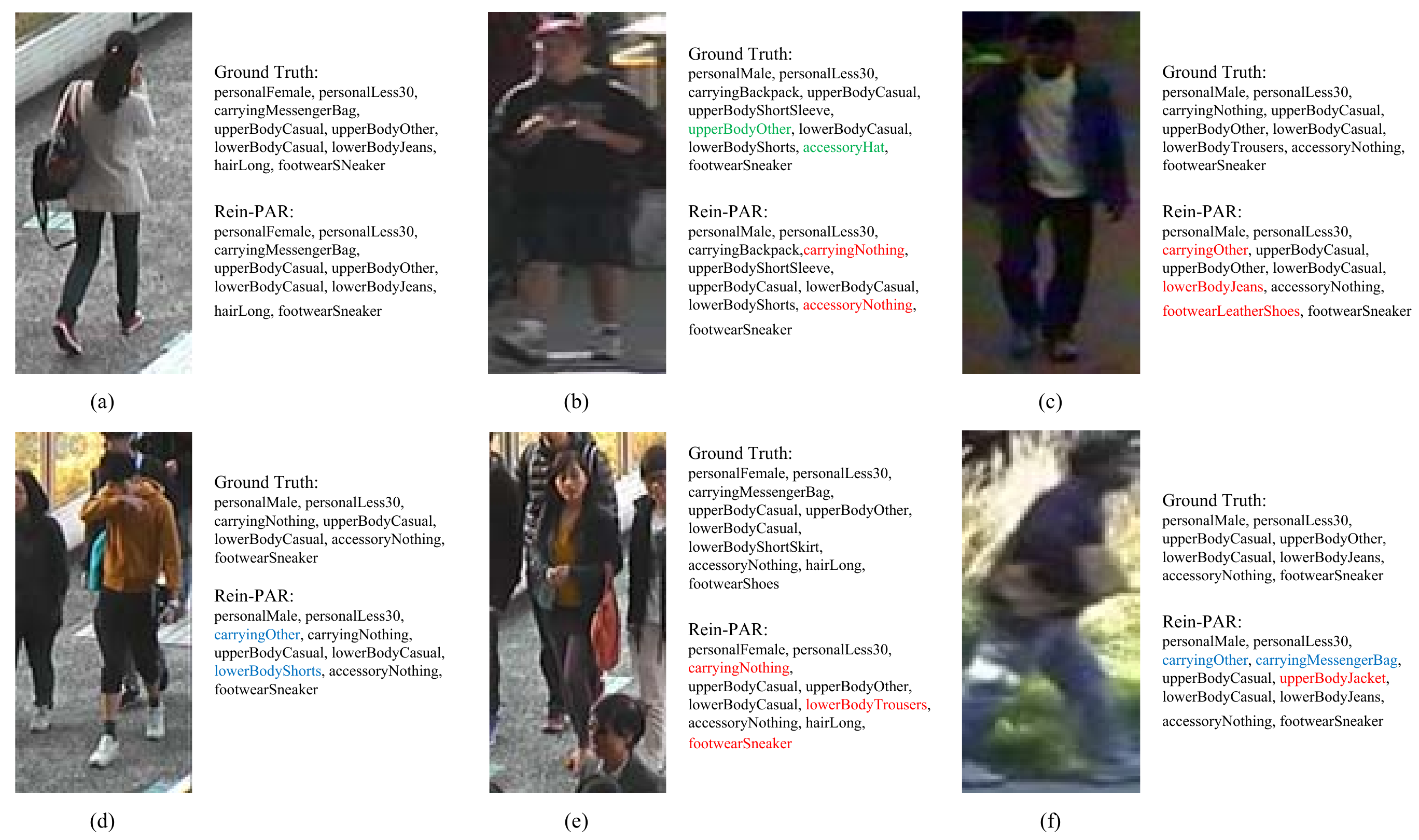}
		\end{center}
		
		\caption{Pedestrian attribute recognition results by Rein-PAR on the PETA dataset, where the black font indicates the correct recognition result, the red indicates the incorrectly recognized attribute, the blue is  the redundantly recognized attribute, and the green represents the unrecognized attribute. Among the six pedestrian images, (a) is an example where the recognition results are completely correct, and (b-f) are examples where the recognition results are defective Note that the images are blurred due to the low image resolution in the dataset.}
	\label{fig6}
	\end{figure*}

Fig. \ref{fig6} shows some recognition results on the PETA dataset. It consists of the image, the corresponding Ground Truth, and the recognition results. Firstly, we can observe that the number of attributes of each pedestrian image in the dataset is not necessarily the same. Secondly, it can be observed that the recognition errors are often caused by illumination, low resolution, blur, multiple pedestrian targets, etc. Take Fig. \ref{fig6} (e) as an example, our approach wrongly considers that the female has three attributes, namely ``carryingNothing", ``lowerBodytrousers" and ``footwearSneaker"  due to the interference of surrounding pedestrians. Thirdly, the redundant attributes are often closely related to pedestrian images. For example, the male in Fig. \ref{fig6} (d) may actually has the attributes of ``carryingOther" and ``lowerBodyShorts", although they are not annotated in the ground-truth. It should be noted that these unannotated categories will be regarded as negative categories during training, which will interfere the training stage. 

\subsection{Ablation Studies}
We analyze the effectiveness of each part of our proposed Rein-PAR in Table \ref{table7}. We first consider
the following variants:

\textbf{Baseline} refers to the approach that does not apply the approaches of Attributes Grouping Strategy (AGS) and Group Optimization Reward (GOR).

\textbf{Baseline+AGS} is the approach applying AGS for prediction based on RPAR.

\textbf{Rein-PAR} is the proposed approach, which incorporates both AGS and GOR.

\begin{table}[h]
	
	\caption{\upshape  Ablation studies of Rein-PAR on PETA, RAP and PA100K datasets. The best results are marked in boldface.}
	\centering
	\tabcolsep 12pt
	\renewcommand\arraystretch{1.2}
	\begin{tabular}{ll|lllll}
	\hline
	\multicolumn{1}{l|}{Dataset}                 & \multicolumn{1}{c|}{Method} & \multicolumn{1}{c}{mA} & \multicolumn{1}{c}{Acc} & \multicolumn{1}{c}{Prec} & \multicolumn{1}{c}{Rec} & \multicolumn{1}{c}{F1} \\ \hline
	\multicolumn{1}{l|}{\multirow{3}{*}{PETA}}   & Baseline                        & 80.06                  & 72.63                   & 84.28                    & 81.54                   & 82.20                  \\
	\multicolumn{1}{l|}{}                        & Baseline+AGS               & 83.14                  & 77.86                   & \textbf{86.02}           & 85.55                   & 85.49                  \\
	\multicolumn{1}{l|}{}                        & Rein-PAR                    & \textbf{85.51}         & \textbf{78.45}          & 84.08                    & \textbf{88.77}          & \textbf{85.91}         \\ \hline
	\multicolumn{1}{l|}{\multirow{3}{*}{RAP}}    & Baseline                        & 69.90                  & 63.71                   & 79.53                    & 75.05                   & 76.35                  \\
	\multicolumn{1}{l|}{}                        & Baseline+AGS               & 75.68                  & \textbf{67.26}          & \textbf{80.94}           & 78.02                   & \textbf{79.03}         \\
	\multicolumn{1}{l|}{}                        & Rein-PAR                    & \textbf{81.67}         & 66.24                   & 73.24                    & \textbf{85.80}          & 78.68                  \\ \hline
	\multicolumn{1}{l|}{\multirow{3}{*}{PA100K}} & Baseline                       & 74.66                  & 74.49                   & 85.78                    & 82.87                   & 83.72                  \\
	\multicolumn{1}{l|}{}                        & Baseline+AGS               & 78.56                  & 76.41                   & \textbf{86.63}           & 84.38                   & 85.04                  \\
	\multicolumn{1}{l|}{}                        & Rein-PAR                    & \textbf{80.55}         & \textbf{77.20}          & 84.76                    & \textbf{87.67}          & \textbf{85.70}         \\ \hline
	\end{tabular}
	\label{table7}
	\end{table}

It can be observed that even the Baseline approach could achieve satisfactory results. For example, it is better than GSR-MAR and ACN on the RAP dataset, and also better than DeepMAR, HP-Net and GSR-MAR on the PA100K dataset. Furthermore, it is observed that the Baseline+AGS approach outperforms the Baseline by 3.08\% on PETA, 5.78\% on RAP and 3.9\% on PA100K on mA metric. And on the F1 metric, the Baseline+AGS approach has improvements of 3.29\% on PETA, 2.68\% on RAP and 1.32\% on PA100K compared with Baseline. It proves the effectiveness of the grouping attributes strategy. Additionally, Rein-PAR further applies group optimization reward to alleviate the intra-attribute imbalance problem, which brings improvements of 2.27\%, 5.99\%, and 1.99\% on mA on the three datasets compared with Baseline+AGS. In addition to the significant improvement under mA metric, other metrics such as Acc and Rec have also been significantly improved. For instance, the Acc rises from 77.86\% to 78.45\% on PETA. It’s worth noting that the utilization of GOR reduces attention on negative samples, resulting in a decline on the Prec metric.

\subsection{Impact of $\rho$ in the Reward Function} 
It is well-known that different reward functions have different impacts on performance. We select two groups of attributes on the PETA and RAP datasets and test the impact of different values of $\rho$ in the reward function. Table \ref{table8} shows the two groups of attributes and the corresponding attribute ratios.

\begin{table}[h]
	
	\caption{\upshape  Two selected groups of attributes in the PETA and RAP datasets and their corresponding attribute ratios.}
	\centering
	\tabcolsep 10pt
	\renewcommand\arraystretch{1.2}
	\begin{tabular}{lcccccc}
	\cline{1-6}
	\multicolumn{1}{l|}{\multirow{2}{*}{PETA}} & \multicolumn{1}{c|}{Attribute} & LeatherShoes & Sandals & Shoes      & Sneaker     &  \\ \cline{2-6}
	\multicolumn{1}{l|}{}                      & \multicolumn{1}{c|}{Ratio}     & 0.296        & 0.02    & 0.363      & 0.216       &  \\ \cline{1-6}
	\multicolumn{1}{c|}{\multirow{4}{*}{RAP}}  & \multicolumn{1}{c|}{Attribute} & Calling      & Talking & Gathering  & Holding     &  \\ \cline{2-6}
	\multicolumn{1}{c|}{}                      & \multicolumn{1}{c|}{Ratio}     & 0.034        & 0.032   & 0.092      & 0.025       &  \\ \cline{2-6}
	\multicolumn{1}{c|}{}                      & \multicolumn{1}{c|}{Attribute} & Pushing      & Pulling & CarrybyArm & CarrybyHand &  \\ \cline{2-6}
	\multicolumn{1}{c|}{}                      & \multicolumn{1}{c|}{Ratio}     & 0.01         & 0.018   & 0.023      & 0.129       &  \\ \cline{1-6}

	\end{tabular}
	\label{table8}
	\end{table}

\begin{figure}[htbp]
	\quad
	\centering
	\subfigure[on PETA dataset]{
	\includegraphics[width=9cm]{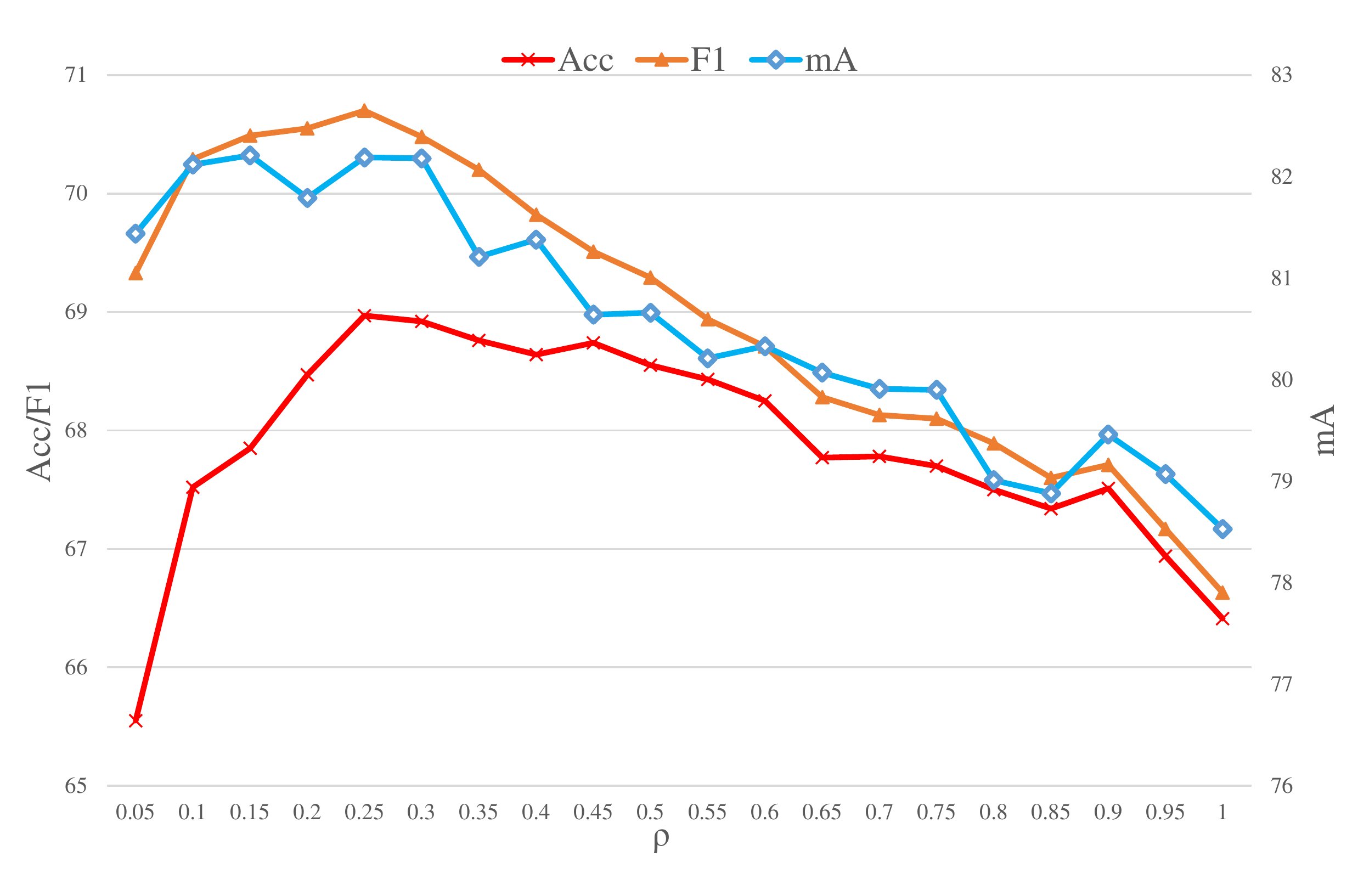}
	}
	\quad
	\subfigure[on RAP dataset]{
	\includegraphics[width=9cm]{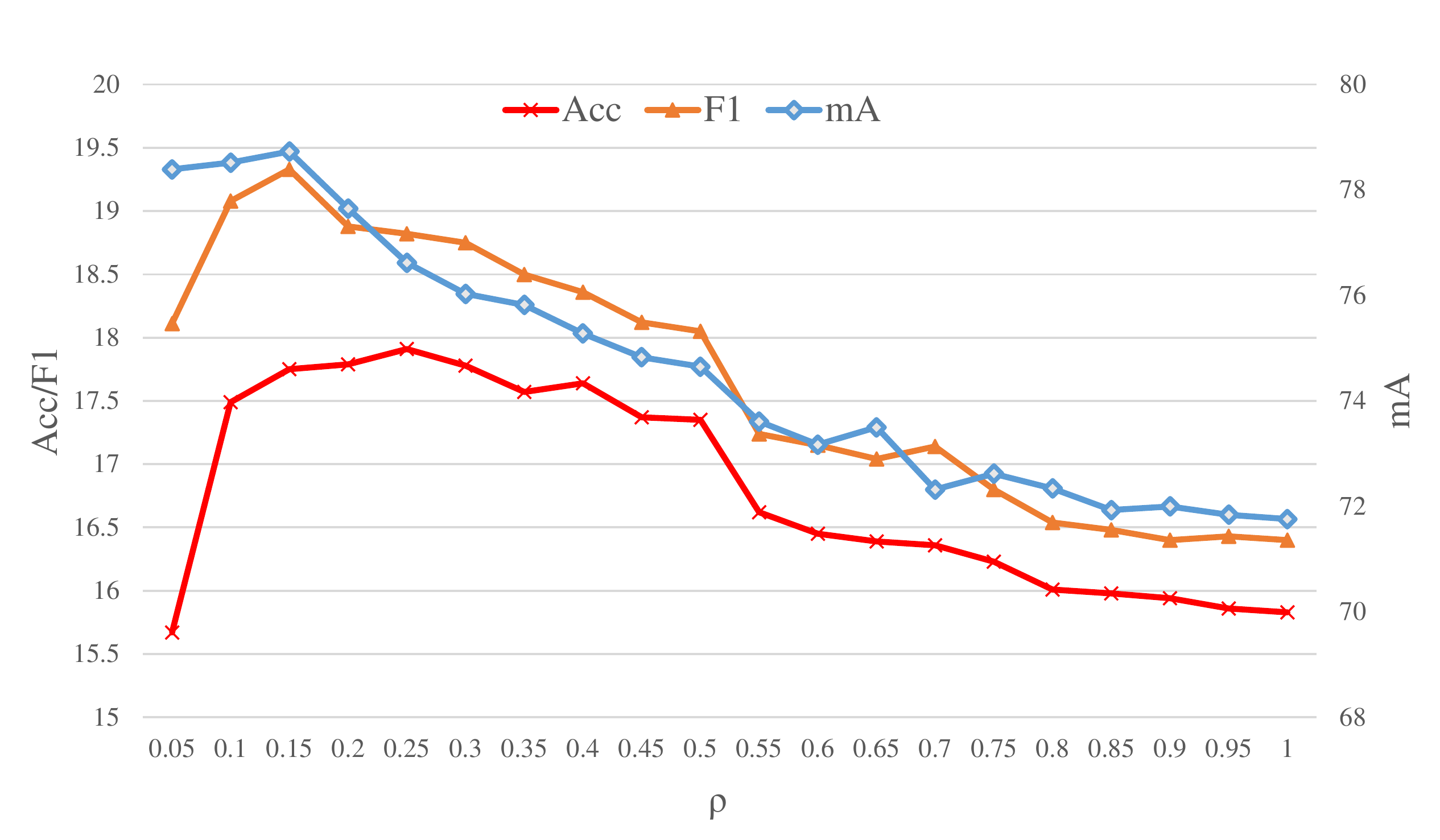}
	}
	\caption{The impact of $\rho$ on Acc, F1 and mA. }
\label{fig7}
\end{figure}

We test the performance when the $\rho$ is from 0.05 to 1, with the interval is 0.05. The impact of $\rho$ is visualized in terms of Acc, F1 and mA, as shown in Fig. \ref{fig7}. It should be noted that, because the difficulty of recognizing each group of attributes is different, some of the metrics in some groups are lower. For example, as shown in Fig. 7(b), the attribute group selected from the RAP dataset are mostly uncommon attributes, which are more difficult to recognize, so the Acc and F1 metrics are low. We can observe that value of  $\rho$ has a significant effect on performance, and the optimal $\rho$ values corresponding to different groups are different. We could also observe that the influence trends of $\rho$ on the three metrics of Acc, F1 and mA are almost the same. The value of the appropriate $\rho$ appears at the peak of the evaluation metrics curves.
\section{Conclusion} 
This paper has proposed the Rein-PAR approach for addressing the PAR task. Different from previous approaches, Rein-PAR defines PAR as a Markov decision-making process for the first time, and employs the Deep Q-learning algorithm to train the network to recognize attributes. Moreover, an attribute grouping strategy is applied to alleviate inter-attribute imbalance problem, and a group optimization reward function is further developed to alleviate the intra-attribute imbalance problem. Experimental results on PETA, RAP and PA100K datasets have demonstrated the effectiveness of our approach. Rein-PAR is a successful attempt to apply reinforcement learning on PAR, which demonstrates that reinforcement learning has considerable potential in PAR. In the future, we consider to construct a more appropriate Markov decision process, a reward function that is more in line with the actual situation of PAR task, and utilize more advanced reinforcement learning algorithms for training.

\bibliographystyle{unsrt}  
\bibliography{rein-par-refs}

\end{document}